\documentclass[11pt]{article}

\usepackage[preprint]{acl}

\usepackage{times}
\usepackage{latexsym}
\usepackage{graphicx}
\usepackage{subcaption}  
\usepackage{url}

\usepackage{pgfplots}
\pgfplotsset{compat=1.18}

\usepackage[T1]{fontenc}

\usepackage[utf8]{inputenc}

\usepackage{microtype}

\usepackage{inconsolata}

\usepackage{graphicx}

\usepackage{amsmath}
\usepackage{xspace}
\usepackage{cleveref}
\usepackage{booktabs}
\newcommand{\term}[1]{\texttt{\small #1}}

\title{A Joint Multitask Model for Morpho-Syntactic Parsing}

\author{%
  Demian Inostroza, Mel Mistica, Ekaterina Vylomova, Chris Guest, Kemal Kurniawan\\
  University of Melbourne\\
  \parbox{\linewidth}{\centering\ttfamily
    \{inostrozaad, misticam, ekaterina.vylomova,\\%
    chris.guest, kurniawan.k\}@unimelb.edu.au
  }
}

\newcommand{\wordtypeclf}{content word identification\xspace}
\newcommand{\wordtypeclfshort}{CWI\xspace}

\begin{document}
\maketitle
\begin{abstract}

We present a joint multitask model for the UniDive 2025 Morpho-Syntactic Parsing shared task, where systems predict both morphological and syntactic analyses following novel UD annotation scheme.
Our system uses a shared XLM-RoBERTa encoder with three specialized decoders for \wordtypeclf, dependency parsing, and morphosyntactic feature prediction.
Our model achieves the best overall performance on the shared task's leaderboard covering nine typologically diverse languages, with an average MSLAS score of 78.7\%, LAS of 80.1\%, and Feats F1 of 90.3\%.
Our ablation studies show that matching the task's gold tokenization and \wordtypeclf are crucial to model performance. Error analysis reveals that our model struggles with core grammatical cases (particularly \term{Nom-Acc}) and nominal features across languages.\footnote{Our code and models are publicly available: \url{https://github.com/DemianInostrozaAmestica/shared_task_UD_official}}

\end{abstract}

\section{Introduction}\label{sec:intro}

The UniDive 2025 Morpho-Syntactic Parsing shared task \citep{goldman2025unidive} introduces a novel framework for dependency parsing that seeks to bridge the traditional divide between morphological and syntactic analysis.
In conventional Universal Dependencies~\cite{nivre-etal-2020-universal}, morphology and syntax are treated as distinct modules operating at different linguistic levels, with word boundaries serving as the interface between them. However, this separation has led to significant inconsistencies in how different languages and even different treebanks for the same language handle word segmentation and grammatical analysis. The shared task proposes to address these long-standing challenges by reorganizing grammatical representation around the content-function distinction rather than relying on theoretically problematic word boundaries, proposing a more typologically consistent approach to multi-linguistic parsing. For instance, in the sentence `From the AP comes this story' shown in Table \ref{tab:new-annotation}, traditional UD treats `From' as a dependent of `AP' with the deprel \term{case}, while the new framework transfers the grammatical meaning of `From' as a morphosyntactic feature \term{Case=Abl} (Ablative) directly onto the content word `AP'.

\begin{table}
  \centering
  \footnotesize
  \begin{tabular}{clp{2.5cm}cc}
    \textbf{ID} & \textbf{Token} & \textbf{FEATS} & \textbf{HEAD} & \textbf{DEPREL} \\
    1 & From & \_ & \_ & \_ \\
    2 & the & \_ & \_ & \_ \\
    3 & AP & Case=Abl|Definite=Def| Number=Sing & 4 & obl \\
    4 & comes & Mood=Ind|Polarity=Pos| Tense=Pres|VerbForm=Fin| Voice=Act & 0 & root \\
    5 & this & Number=Sing| PronType=Dem & 6 & det \\
    6 & story & Number=Sing & 4 & nsubj \\
    7 & : & \_ & \_ & \_
  \end{tabular}
  \caption{Example of the new annotation scheme used in the shared task}\label{tab:new-annotation}
\end{table}

\begin{figure}[t]
  \centering
  \includegraphics[width=\columnwidth]{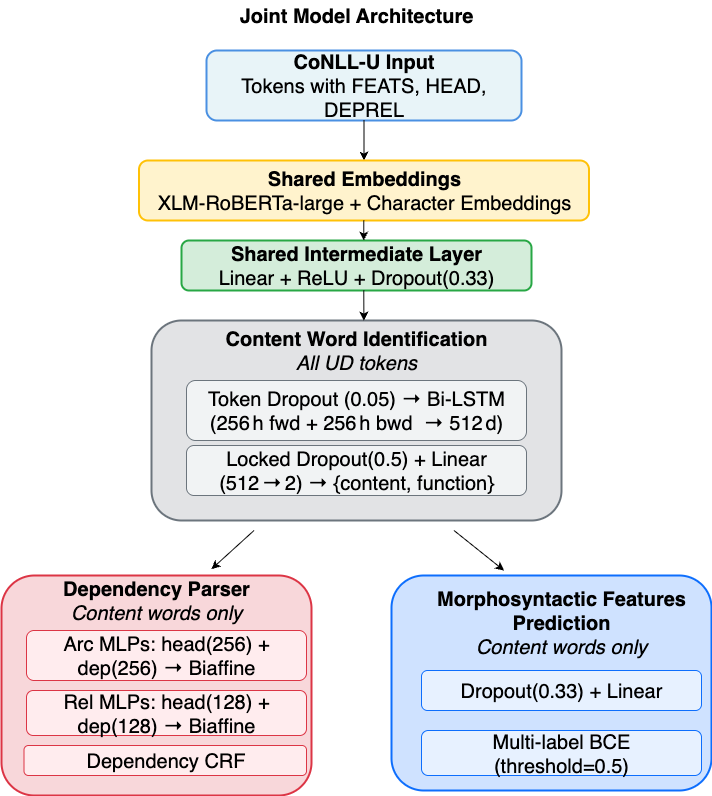}
  \caption{Joint model architecture for the shared task.}
  \label{fig:model-architecture}
\end{figure}

The task requires systems to predict both labeled dependency arcs and morphosyntactic features, but with a difference from standard Universal Dependencies parsing: the dependency tree consists only of content words (lexical words carrying semantic meaning like nouns, verbs, and adjectives), while function words (grammatical elements like adpositions, articles, and auxiliaries) contribute their grammatical information as features on related content words. 

While the content-function distinction is explicit in the training data, systems must identify this distinction themselves at test time from raw text. This identification determines which words participate in the dependency tree and which contribute features to other words. Additionally, the multi-label nature of features, where a content word can have multiple feature values for a given feature class, like \term{Case=Ine;Atr},\footnote{Ine=Inessive, “inside an enclosed area”; Atr=“complement, attribute”. Both definitions come from the official \term{Case} inventory supplied by the shared-task organisers.} requires models to learn intricate morphosyntactic patterns.

We present a joint multitask model~(\Cref{fig:model-architecture}) that explicitly addresses these challenges through three specialized decoders sharing a common XLM-RoBERTa encoder \cite{conneau-etal-2020-unsupervised}, initialized from pre-trained multilingual representations. We design \wordtypeclf as an explicit task to be learned by the model rather than relying on intuition-driven heuristics. We participate in the multilingual track, training separate models for each of the nine languages, allowing us to tune hyperparameters specifically for each language's characteristics while still benefiting from multilingual pretrained representations. On the shared task's results, our system achieves the best overall performance with average scores of 78.7\% MSLAS, 80.1\% LAS, and 90.3\% Feats F1 across all languages. Additionally, our model ranks first on each individual language, demonstrating the effectiveness of multitask learning for this task.

Our error analysis yields three main observations: (1) errors in tokenization and \wordtypeclf cascade through the pipeline, with gold annotations improving MSLAS by up to 12 points; (2) the majority of residual errors lie in nominal morphology—\term{Gender}, \term{Number}, and \term{Case}—with common \term{Nominative–Accusative} swaps; and (3) syntactic mislabels are concentrated in the \term{nmod} versus \term{obl} relation.

\section{System Description}\label{sec:system}

\subsection{Model Overview}

We propose a joint multitask model implemented using the Flair framework \cite{akbik2019flair} for morphosyntactic parsing, as shown in \Cref{fig:model-architecture}. Although the evaluation metrics assess only dependency arcs and morphosyntactic features, producing these outputs requires distinguishing between content and function words. Because this classification is not given at test time, we treat it as an additional prediction task. Our system uses the large version of XLM-RoBERTa augmented with character embeddings \cite{akbik2018coling} as a shared encoder, both provided by the Flair framework. This encoder's output is then passed through a shared intermediate layer (linear transformation with ReLU and dropout) before being fed to three specialized decoders: \wordtypeclf, morphosyntactic feature prediction, and dependency parsing.

\subsection{Decoders}

\paragraph{Content word identification.}
The \wordtypeclf decoder
accepts tokens as input.
Each token’s contextual embedding computed by the shared intermediate layer is passed through a bidirectional LSTM (256 hidden units in both directions).
The LSTM output is then passed through a linear layer with 2 output units, each corresponds to “content” vs. “function” respectively.
Training uses two forms of regularisation: token‑level~(word) dropout---zeroing the entire embedding of 5\% of UD tokens---and locked dropout that masks 50\% of the LSTM outputs with the same pattern across all timesteps. Class‑weighted cross‑entropy loss function is then used to compensate for the imbalance between the number of content and function tokens.

\paragraph{Morphosyntactic features.}
The morphosyntactic features decoder consists of a single linear layer that performs multi-label classification directly from the output of the shared intermediate layer. For each content word, it outputs probabilities for all possible feature-value pairs in the vocabulary~(e.g., \term{Case=Gen}, \term{Number=Sing}, \term{Voice=Act}). Using sigmoid activation with a 0.5 threshold, the model can predict multiple features per token—for instance, a noun might simultaneously have \term{Number=Plur} and \term{Case=Gen}. Complex features with multiple values (like \term{Case=Ine;Atr}) are handled by predicting each component separately, allowing the model to learn different value combinations. Function words bypass this decoder entirely and receive `\_' as their feature value. At training time, we use gold content word ~(i.e. checking if its feature values exist). In contrast, we use the predicted content words by the \wordtypeclf at test time.

\paragraph{Dependency parser.}
The parsing decoder employs separate multilayer perceptrons (MLPs) for arc and relation prediction with biaffine attention mechanisms, following \citet{DBLP:journals/corr/DozatM16}.
The arc MLPs have 256 hidden units while the relation MLPs use 128 units, both with layer normalization and ReLU activation. Operating exclusively on content words, we frame the parser as a conditional random field over projective dependency trees that we implement using TorchStruct \cite{alex2020torchstruct}.
Similar to the morphosyntactic feature decoder, we use gold and predicted content word at training and test time respectively.

\subsection{Data Handling and Inference}

While the shared task data includes abstract nodes for representing implicit arguments, we initially attempted to handle them through sequence labeling by inserting mask tokens at potential abstract node positions. However, this approach introduced noise that degraded performance across all metrics, as incorrect abstract node predictions propagated errors to downstream decoders. Therefore, our final system filters out abstract nodes during data loading, simplifying the parsing task while improving overall performance.

During inference, raw text is first segmented into word tokens using Stanza \cite{qi2020stanza}. Since tokenization quality impacts downstream performance but is not the focus of this shared task, we choose to leverage Stanza's pre-trained models rather than training custom tokenizers. For each language, we evaluated different Stanza model variants on the development set and selected those that best matched the gold tokenization (e.g., HTB for Hebrew, IMST for Turkish). This selection was done manually by running the full pipeline with each available Stanza model variant and choosing the one that achieved the highest metrics on the official evaluation script.

We apply minimal post-processing to ensure valid output. For \wordtypeclf, tokens with confidence below 0.6 that appear between two tokens of the opposite type are relabeled to match their context (e.g., a low-confidence function word between two content words becomes content). As a fallback for extreme cases where \wordtypeclf predicts all tokens as function words (particularly in very short sentences of 2-3 tokens), we force the first token to be content with \term{deprel=`root'} and \term{features=`|'}. This ensures every sentence has at least one parseable token.

\subsection{Training Objective and Optimization}

The model is trained end-to-end using a weighted sum of the three decoders' losses: $\mathcal{L}_{\text{total}} = w_{\text{parser}}\mathcal{L}_{\text{parser}} + w_{\text{morph}}\mathcal{L}_{\text{morph}} + w_{\text{CWI}}\mathcal{L}_{\text{CWI}}$, where the weights are hyperparameters tuned for each language. The parser uses negative log-likelihood loss over projective trees, the morphosyntactic decoder uses binary cross-entropy for multi-label classification, and the \wordtypeclf uses class-weighted cross-entropy to handle class imbalance.

\section{Experimental Setup}\label{sec:setup}

The shared task provided training and development sets for multiple languages. To simulate a realistic evaluation scenario, we split the official training data into 90\% for training and 10\% for development, using the official development set as our local test set. This allowed us to tune hyperparameters and select models before the official test release. The languages included in our experiments were English, Turkish, Hebrew, Czech, Polish, Portuguese, Italian, Serbian, and Swedish, with training sizes ranging from approximately 3,000 to 10,000 sentences depending on the language.

We develop a custom data loader to handle the modified CoNLL-U format used in the shared task. The loader automatically extracts content words by examining the FEATS column, where `\_' indicates function words and any other value indicates content words. As mentioned before, we filter out abstract nodes during loading.

All models are trained using AdamW optimizer~\cite{loshchilov2019decoupled} with an initial learning rate of $2\times10^{-5}$ and batch size of 16 for 25 epochs. We employ early stopping with patience of 1 epoch and learning rate reduction by factor 0.5 when validation loss plateaus. Training is performed on a NVIDIA A100 GPU with 32GB RAM on a high-performance computing cluster, with each model taking approximately 1-5 hours to converge.

We perform grid search over task-specific loss weights on our development split. The optimal weights varied by language—for example, Turkish benefited from weighting parsing and morphological feature losses twice as much as content word identification (2.0:2.0:1.5), while English performed better with parsing weighted most heavily, followed by morphological features and content word identification (2.0:1.5:1.0).

For each language, we train three models with different random seeds using the same hyperparameter configuration to verify training stability and robustness. All three models are evaluated on our local test set (the official development set) using the shared task's official evaluation script.

Once hyperparameters are selected, we retrain a single model for each language using the complete official training and development data combined. These final models use the same hyperparameters determined during development. These models are used to generate predictions on the official covered test set, which contains only raw text without annotations. Evaluation is performed using the official script which computes three metrics: MSLAS (morphosyntactic features F1 only on correctly parsed tokens), LAS (labeled attachment score), and Feats F1 (morphosyntactic features F1).

\section{Results}\label{sec:results}

This section is divided into two parts: first, we present official test results from models trained on all available data (official train + dev combined) and evaluated on the covered test set; second, we report development results using our local data splits (90\% train, 10\% dev, official dev as test) to analyze design choices and hyperparameter impact.

\subsection{Official Test Results}\label{sec:test_results}

\begin{table}\small
  \centering
  \begin{tabular}{@{}lccc@{}}
    \toprule
    \textbf{Language} & \textbf{MSLAS} & \textbf{LAS} & \textbf{Feats} \\
    \midrule
    Czech       & 87.1 & 88.0 & 95.2 \\
    English     & 83.8 & 85.1 & 94.9 \\
    Hebrew      & 68.7 & 71.4 & 83.4 \\
    Italian     & 73.0 & 73.7 & 84.7 \\
    Polish      & 75.0 & 76.5 & 86.2 \\
    Portuguese  & \textbf{88.9} & \textbf{89.5} & 94.8 \\
    Serbian     & 86.6 & 88.3 & 95.6 \\
    Swedish     & 86.6 & 87.7 & \textbf{95.7} \\
    Turkish     & 58.7 & 60.9 & 82.1 \\
    \midrule
    \textbf{Average} & 78.7 & 80.1 & 90.3 \\
    \bottomrule
  \end{tabular}
  \caption{Official test results on the covered test set. Our system achieved the highest average MSLAS score (78.7\%) among all submissions.}
  \label{tab:official_results}
\end{table}

\Cref{tab:official_results} presents the official test results from models trained on all available data. Our system achieved the highest performance among all submissions with an average MSLAS of 78.7\%. The results show strong performance across most languages, with MSLAS scores exceeding 83\% for seven of the nine languages. Portuguese (88.9\%) and Czech (87.1\%) achieved the highest scores, consistent with our development results. The morphologically complex languages continued to present challenges—Turkish (58.7\%) and Hebrew (68.7\%) showed the lowest performance.

\begin{table}\small
  \centering
  \begin{tabular}{@{}lccc@{}}
    \toprule
    \textbf{System} & \textbf{MSLAS} & \textbf{LAS} & \textbf{Feats} \\
    \midrule
    Our model           & \textbf{78.7} & \textbf{80.1} & \textbf{90.3} \\
    \midrule
    baseline\_multi     & 47.3 & 55.4 & 64.2 \\
    baseline\_cross     & 36.7 & 51.2 & 50.6 \\
    baseline\_finetune  & 33.0 & 36.1 & 52.3 \\
    \bottomrule

  \end{tabular}
  \caption{Comparison with baseline systems (average across all languages).}
  \label{tab:baseline_comparison}
\end{table}

The baseline systems provide important context for understanding the task's difficulty (\Cref{tab:baseline_comparison}). The multilingual few-shot baseline achieved moderate performance (average MSLAS 47.3\%), while the cross-lingual few-shot approach struggled significantly (36.7\%), highlighting the importance of language-specific examples. The finetuned BERT baseline performed poorest (33.0\%), suggesting that the reformulated parsing task with its content-function distinction and expanded feature inventory benefits from specialized modeling approaches. Our 31.4 point improvement over the best baseline (78.7\% vs 47.3\%) indicates that combining pretrained representations with task-specific architectural components can effectively address the challenges of unified morphosyntactic parsing.

\subsection{Development Results}\label{sec:dev_results}

\begin{figure}[t]
\centering
\begin{tikzpicture}
\begin{axis}[
  width=\columnwidth, height=0.5\columnwidth,
  ymin=50, ymax=100,
  enlarge x limits=0.2,
  xtick={1,2,3}, xticklabels={Full, GoldTok, GoldWT},
  ymajorgrids=true,
  tick label style={font=\footnotesize},
  ylabel=MSLAS,
  legend to name=mslaslegend,
  legend columns=2,
  legend style={draw=none,font=\scriptsize,/tikz/every even column/.style={column sep=8pt}},
  every axis plot/.append style={mark=*, mark size=1.8pt, line width=0.9pt}
]

\addplot coordinates {(1,75.2)(2,84.5)(3,85.7)}; \addlegendentry{Hebrew}
\addplot coordinates {(1,85.1)(2,87.1)(3,89.2)}; \addlegendentry{Italian}
\addplot coordinates {(1,84.1)(2,84.1)(3,87.2)}; \addlegendentry{Swedish}
\addplot coordinates {(1,53.9)(2,54.6)(3,56.8)}; \addlegendentry{Turkish}

\end{axis}
\end{tikzpicture}

\pgfplotslegendfromname{mslaslegend}
\caption{MSLAS across setups (Full, GoldTok, GoldWT) for four languages with the largest gains. 
\textbf{Full}: predicted tokenization and predicted content word identity.
\textbf{GoldTok}: gold tokenization with predicted content word identity.
\textbf{GoldWT}: gold tokenization plus gold content word identity.}
\label{fig:ablation_mslas_top4}
\end{figure}
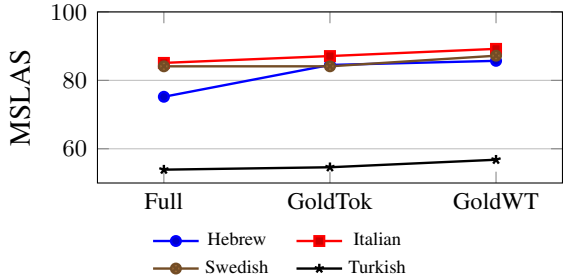

The ablations in \Cref{fig:ablation_mslas_top4} show that most of the gain comes from using gold tokenization, with a smaller but consistent boost from explicit content/function labeling. Hebrew makes this clear: MSLAS goes from 75.2 (\textbf{Full}) → 84.5 (\textbf{GoldTok}, +9.3) → 85.7 (\textbf{GoldWT}, +1.2; +10.5 total). This motivates per-language tokenizer selection and modeling \wordtypeclf as a dedicated task.

Loss-weight tuning largely favored \texttt{parser=2.0}, \texttt{morph=1.5}, \texttt{CWI=1.0}; Turkish and Swedish benefited from higher weights on \texttt{morph}/\texttt{CWI} (\Cref{tab:loss_weights}).

\begin{table}\small
  \centering
  \begin{tabular}{@{}lccc@{}}
    \toprule
    \textbf{Language} & \textbf{Parser} & \textbf{Morph} & \textbf{\wordtypeclfshort} \\
    \midrule
    Czech      & 2.0 & 1.5 & 1.0 \\
    English    & 2.0 & 1.5 & 1.0 \\
    Hebrew     & 2.0 & 1.5 & 1.0 \\
    Italian    & 2.0 & 1.5 & 1.0 \\
    Polish     & 2.0 & 1.5 & 1.0 \\
    Portuguese & 2.0 & 1.5 & 1.0 \\
    Serbian    & 2.0 & 1.5 & 1.0 \\
    \midrule
    Swedish    & 2.0 & 1.5 & 1.5 \\
    \midrule
    Turkish    & 2.0 & 2.0 & 1.5 \\
    \bottomrule
  \end{tabular}
  \caption{Optimal loss weight configurations by language. \wordtypeclfshort=\wordtypeclf.}
  \label{tab:loss_weights}
\end{table}

\section{Error Analysis}\label{sec:error}

We performed error analysis on the models trained with our local data splits (90\% train, 10\% dev, official dev as test). We analyzed only the first seed model for each language, as the low standard deviations indicate minimal variation across seeds. The analysis uses scripts that replicate the official evaluation logic to ensure our error categorization matches the scoring methodology.

\subsection{Nominal Morphology Errors}

The main feature prediction errors occur in nominal morphology, with Gender, Number, and Case showing the highest confusion rates. Since languages have different feature inventories (e.g., Czech includes Dual while others do not), creating a unified confusion matrix is not feasible. We selected Czech as a representative example because it has by far the most training data points, resulting in more stable model behavior.

Our analysis of Czech reveals strong overall performance, with 99.1\% accuracy for Gender and 99.6\% for Number predictions. For Gender, the model correctly classifies the vast majority of instances, with Feminine (12,405 correct), Masculine (14,914 correct), and Neuter (5,325 correct) all showing high diagonal values in the confusion matrix. The annotation scheme includes syncretic forms like "Fem,Masc" for grammatically ambiguous cases. The most common confusions occur between Masculine and Feminine (110 instances misclassified as Feminine when Masculine was correct), though these remain relatively rare. Similarly, for Number, Singular (24,653 correct) and Plural~(9,587 correct) are accurately predicted, with minimal confusion between categories (only 44 Singular instances misclassified as Plural, and 85 Plural instances misclassified as Singular).

Since our model uses multi-label classification with sigmoid activation (threshold 0.5), it occasionally predicts semantically incompatible feature combinations—for instance, simultaneously predicting both a specific gender value (e.g., "Fem") and a syncretic form containing that value (e.g., "Fem,Masc"). While these semantically nonsensical predictions are rare (occurring in fewer than 100 instances out of over 30,000), they suggest that post-processing constraints based on linguistic compatibility rules could eliminate such predictions and further improve the performance.

For \term{Case} features, plotting a confusion matrix is impractical due to the >100 possible values in the expanded inventory. While there is some variation across languages, aggregating the most frequent errors reveals consistent patterns. \Cref{tab:case_top10_errors} shows the 10 most common \term{Case} confusions averaged across all languages. The high frequency of \term{Nom-Acc} confusions (154 and 140 instances) reflects both the prevalence of these cases in the data and their potential ambiguity—distinguishing core arguments becomes particularly challenging in complex sentences with long-distance dependencies or multi-clause structures. This pattern holds across languages despite their individual variations, suggesting that even within the expanded \term{Case} system, these fundamental grammatical distinctions remain challenging when syntactic complexity increases. These systematic errors in core grammatical cases suggest a targeted improvement strategy: increasing loss weights for frequently confused cases (especially \term{Nom/Acc}) during training. Given our joint model architecture where all tasks share embeddings, better representation of these central arguments could benefit dependency parsing as well.

\begin{table}\small
  \centering
  \begin{tabular}{@{}rll@{}}
    \toprule
    \textbf{Count} & \textbf{Gold case} & \textbf{Predicted case} \\
    \midrule
    154 & Acc        & Nom        \\
    140 & Nom        & Acc        \\
    77  & Nom        & Conj;Nom   \\
    47  & Nom        & Gen        \\
    44  & Conj;Nom   & Nom        \\
    43  & Gen        & Nom        \\
    36  & Acc        & Gen        \\
    25  & Gen        & Acc        \\
    22  & Gen        & Conj;Gen   \\
    15  & Dat        & Ins        \\
    \bottomrule
  \end{tabular}
  \caption{Top 10 most frequent case prediction errors (average across all languages).}
  \label{tab:case_top10_errors}
\end{table}

\subsection{Spatial Case Results}

We evaluate our model's performance on the fine-grained spatial \term{Case} values, a particularly challenging subset due to the numerous possible inflectional meanings that this domain contains.\footnote{These notions are understood as defined by \citet{haspelmath2025grammatical}: “inflectional meaning” designates the specific meaning conveyed by an inflected form (for example, ablative), and “inflectional domain” denotes the broader class of related properties in which this meaning is categorized (for example, case).} The complete inventory of spatial cases includes over 40 fine-grained distinctions. \Cref{tab:spatial-cases} shows high performance across all languages (F1 scores 89.2-98.7\%), demonstrating that our model successfully learned the unified \term{Case} system for spatial meanings. This annotation scheme directly names inflectional meanings regardless of the grammatical markers used - for instance, in Polish, when ablative meaning is expressed periphrastically through a clitic (an adposition)\footnote{The classification of adpositions as clitics follows the definition proposed by \citet{Haspelmath_2023}.} plus an inflected form (a root with a genitive case affix), the system assigns the inflectional meaning (e.g., \term{Case=Abl}) instead of the genitive meaning conveyed by the suffix on its own. Our model's performance on these distinctions suggests it effectively captures the mapping between diverse surface forms and their underlying spatial semantics. This opens opportunities for injecting linguistic knowledge about spatial relations in downstream applications, leveraging the semantic transparency of the annotation scheme. 

\begin{table}\small
  \centering
  \begin{tabular}{@{}lccc@{}}
    \toprule
    \textbf{Language} & \textbf{Precision} & \textbf{Recall} & \textbf{F1} \\
    \midrule
    Czech & 98.2 & 98.4 & 98.3 \\
    English & 93.3 & 90.3 & 91.8 \\
    Hebrew & 88.4 & 90.0 & 89.2 \\
    Italian & 98.0 & 97.0 & 97.5 \\
    Polish & 98.4 & 97.2 & 97.8 \\
    Portuguese & \textbf{98.5} & \textbf{99.0} & \textbf{98.7} \\
    Serbian & 96.4 & 93.7 & 95.1 \\
    Swedish & 98.4 & 96.1 & 97.2 \\
    Turkish & 94.7 & 96.4 & 95.6 \\
    \bottomrule
  \end{tabular}
  \caption{Spatial case performance (\%) across languages using micro-averaged metrics.}
  \label{tab:spatial-cases}
\end{table}

\begin{table}\small
  \centering
  \begin{tabular}{@{}ccc@{}}
    \toprule
    \textbf{Count} & \textbf{Gold label} & \textbf{Predicted label} \\
    \midrule
    67 & obl & nmod \\
    63 & nmod & obl \\
    11 & obj & nsubj \\
    11 & advmod & \_ \\
    10 & nmod & flat \\
    10 & nmod & amod \\
    9 & nsubj & obj \\
    9 & iobj & obj \\
    8 & obj & obl \\
    8 & nsubj & root \\
    \bottomrule
  \end{tabular}
  \caption{Top 10 most frequent deprel labeling errors (average across all languages).}
  \label{tab:deprel_top10_confusions}
\end{table}

\begin{figure}[t]
   \centering
   \includegraphics[width=0.45\textwidth]{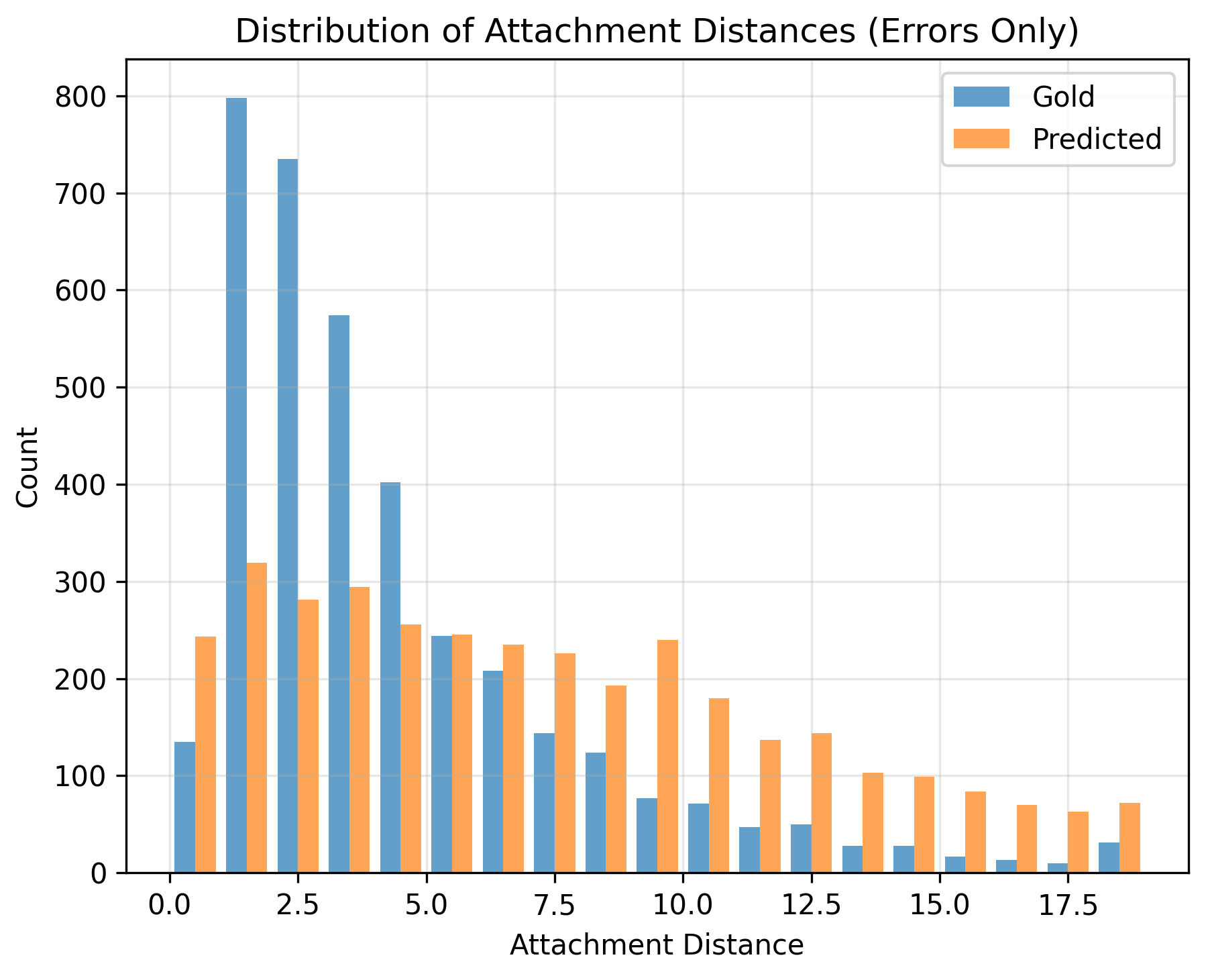}
   \caption{Distribution of attachment distances for parsing errors in Czech.}
   \label{fig:attachment_distances}
\end{figure}

\subsection{Dependency Parsing Errors}

For dependency relation errors, we analyze confusions across all languages since the label inventory is universal. \Cref{tab:deprel_top10_confusions} presents the 10 most frequent labeling errors aggregated across languages. The \term{nmod-obl} confusion dominates with 67 and 63 instances respectively, accounting for over 40\% of the top errors. This pattern is linguistically expected as the boundary between nominal modifiers and oblique arguments could involve borderline cases. 

\begin{figure}
   \centering
   \includegraphics[width=0.45\textwidth]{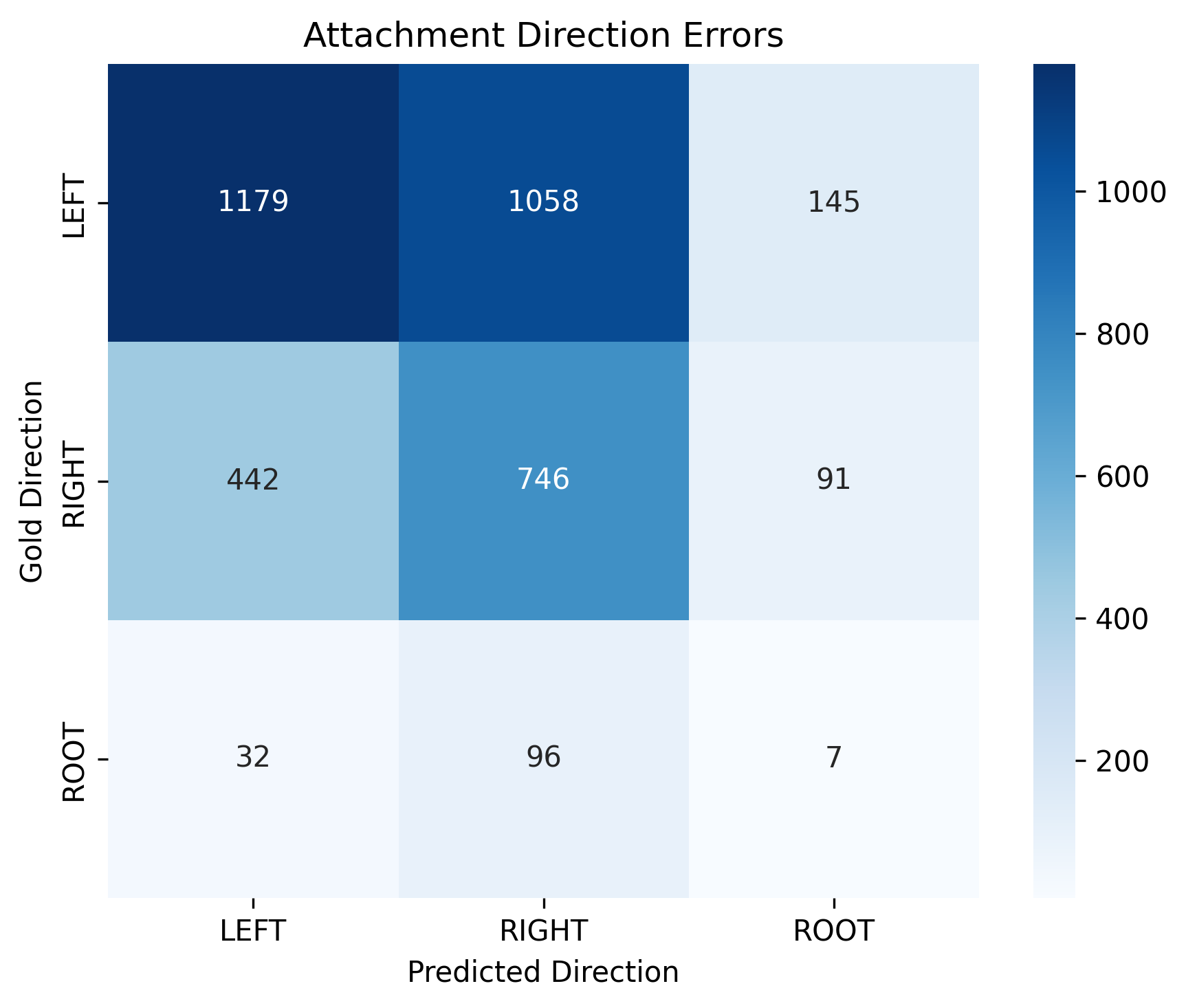}
   \caption{Attachment direction confusion matrix for Czech.}
   \label{fig:attachment_direction}
\end{figure}

Unlike other languages where errors concentrate on the \term{nmod/obl} distinction, Turkish shows a much more dispersed error pattern with confusions spread across many dependency relations. This suggests that our joint architecture may not be optimal for Turkish's non-projective structures and rich morphology. A dedicated non-projective parsing algorithm might better capture Turkish's complex dependency patterns.

Additionally, we analyze attachment distance patterns specifically for parsing errors (i.e., tokens with incorrect head assignments). \Cref{fig:attachment_distances} shows the distribution of attachment distances for Czech parsing errors, comparing gold (blue) versus predicted (orange) distances for these misparsed tokens. The graph reveals that while most gold attachments occur at distances 1-3, the model's errors tend to predict longer distances (note the orange bars extending further right). This indicates the parser frequently overlooks nearby heads in favor of more distant ones when making mistakes. \Cref{fig:attachment_direction} presents a heatmap of misparsed tokens where rows represent gold attachment directions and columns show predicted directions. The strong diagonal (LEFT→LEFT: 1179, RIGHT→RIGHT: 1058) confirms the model correctly identifies attachment direction in most error cases. However, within each correct direction, the parser still selects the wrong head - for instance, when it correctly predicts a leftward attachment, it often chooses a head that is too far to the left.

\section{Conclusions}\label{sec:conclusions}

We present a joint multitask architecture for unified morphosyntactic parsing that achieves first place in the UniDive 2025 shared task. Our key contribution is explicitly modeling \wordtypeclf as a classification task, creating a robust cascade where the identification determines parsing and feature assignment. 

Our analysis reveals systematic error patterns pointing to specific improvement opportunities. \term{Case} confusions concentrate on core grammatical distinctions (\term{Nom-Acc}), while dependency errors reflect the expected challenges at the \term{nmod-obl} boundary. While these patterns are linguistically understandable, they suggest potential room for improvement through weighted training or specialized handling of frequently confused categories, though such optimizations may yield only incremental gains.

A more substantial enhancement to the annotation scheme could be making explicit which function words contribute features to which content words. Currently, function words are marked with `\_' and their grammatical information is incorporated into "related" content words, but these relationships remain implicit. An indexing system could explicitly link each function word to its target content word. This would not only reduce ambiguity in feature assignment but also make the annotation more transparent for researchers unfamiliar with specific languages, as they could trace exactly how morphosyntactic information flows from function words to content words in the unified representation.

Finally, the 30-point performance gap between Portuguese and Turkish highlights fundamental challenges in handling typologically diverse languages within a unified framework. While the parser excels at the predominantly projective structures, Turkish's agglutinative morphology and flexible word order might be introducing some difficulties. The dispersed error patterns observed for Turkish—contrasting with the concentrated confusions in other languages—suggest that the current architecture may not be optimal for highly non-projective languages. Future work could explore specialized parsing algorithms designed for non-projective structures or alternative architectures that better handle long-distance dependencies and flexible word order. Despite these challenges, our results across nine languages demonstrate the viability of joint morphosyntactic modeling for the task.

\section*{Acknowledgments}
We thank the anonymous reviewers for their constructive feedback. This research was supported by The University of Melbourne’s Research Computing Services and the Petascale Campus Initiative. Demian is funded by the Graduate Research Scholarship from the Faculty of Engineering and Information Technology, University of Melbourne. COST (European Cooperation in Science and Technology) kindly provided funding for travel to the event.

\bibliography{custom}

\begin{thebibliography}{11}
\providecommand{\natexlab}[1]{#1}

\bibitem[{Akbik et~al.(2019)Akbik, Bergmann, Blythe, Rasul, Schweter, and Vollgraf}]{akbik2019flair}
Alan Akbik, Tanja Bergmann, Duncan Blythe, Kashif Rasul, Stefan Schweter, and Roland Vollgraf. 2019.
\newblock {FLAIR}: An easy-to-use framework for state-of-the-art {NLP}.
\newblock In \emph{{NAACL} 2019, 2019 Annual Conference of the North American Chapter of the Association for Computational Linguistics (Demonstrations)}, pages 54--59.

\bibitem[{Akbik et~al.(2018)Akbik, Blythe, and Vollgraf}]{akbik2018coling}
Alan Akbik, Duncan Blythe, and Roland Vollgraf. 2018.
\newblock Contextual string embeddings for sequence labeling.
\newblock In \emph{{COLING} 2018, 27th International Conference on Computational Linguistics}, pages 1638--1649.

\bibitem[{Conneau et~al.(2020)Conneau, Khandelwal, Goyal, Chaudhary, Wenzek, Guzm{\'a}n, Grave, Ott, Zettlemoyer, and Stoyanov}]{conneau-etal-2020-unsupervised}
Alexis Conneau, Kartikay Khandelwal, Naman Goyal, Vishrav Chaudhary, Guillaume Wenzek, Francisco Guzm{\'a}n, Edouard Grave, Myle Ott, Luke Zettlemoyer, and Veselin Stoyanov. 2020.
\newblock \href {https://doi.org/10.18653/v1/2020.acl-main.747} {Unsupervised cross-lingual representation learning at scale}.
\newblock In \emph{Proceedings of the 58th Annual Meeting of the Association for Computational Linguistics}, pages 8440--8451, Online. Association for Computational Linguistics.

\bibitem[{Dozat and Manning(2016)}]{DBLP:journals/corr/DozatM16}
Timothy Dozat and Christopher~D. Manning. 2016.
\newblock \href {https://arxiv.org/abs/1611.01734} {Deep biaffine attention for neural dependency parsing}.
\newblock \emph{CoRR}, abs/1611.01734.

\bibitem[{Goldman et~al.(2025)Goldman, Weissweiler, Acar, Alves, Bienati, Eryi{\u g}it, Pagano, Pannitto, Samard{\v z}i{\'c}, Talamo, Wr{\'o}blewska, Zeman, Nivre, and Tsarfaty}]{goldman2025unidive}
Omer Goldman, Leonie Weissweiler, Kutay Acar, Diego Alves, Arianna Bienati, G{\"u}l{\c s}en Eryi{\u g}it, Adriana Pagano, Ludovica Pannitto, Tanja Samard{\v z}i{\'c}, Luigi Talamo, Alina Wr{\'o}blewska, Daniel Zeman, Joakim Nivre, and Reut Tsarfaty. 2025.
\newblock Findings of the {U}ni{D}ive 2025 shared task on multilingual morpho-syntactic parsing.
\newblock In \emph{Proceedings of The {U}ni{D}ive 2025 shared task on multilingual morpho-syntactic parsing}.

\bibitem[{Haspelmath(2023)}]{Haspelmath_2023}
Martin Haspelmath. 2023.
\newblock \href {https://doi.org/10.6092/issn.2785-0943/16057} {Types of clitics in the world’s languages}.
\newblock \emph{Linguistic Typology at the Crossroads}, 3(2):1–59.

\bibitem[{Haspelmath(2025)}]{haspelmath2025grammatical}
Martin Haspelmath. 2025.
\newblock Grammatical markers and inflectional categories: Reconciling the two perspectives.
\newblock Draft, Max Planck Institute for Evolutionary Anthropology, April 9, 2025.

\bibitem[{Loshchilov and Hutter(2019)}]{loshchilov2019decoupled}
Ilya Loshchilov and Frank Hutter. 2019.
\newblock Decoupled weight decay regularization.
\newblock In \emph{International Conference on Learning Representations (ICLR)}.

\bibitem[{Nivre et~al.(2020)Nivre, de~Marneffe, Ginter, Haji{\v{c}}, Manning, Pyysalo, Schuster, Tyers, and Zeman}]{nivre-etal-2020-universal}
Joakim Nivre, Marie-Catherine de~Marneffe, Filip Ginter, Jan Haji{\v{c}}, Christopher~D. Manning, Sampo Pyysalo, Sebastian Schuster, Francis Tyers, and Daniel Zeman. 2020.
\newblock \href {https://aclanthology.org/2020.lrec-1.497/} {{U}niversal {D}ependencies v2: An evergrowing multilingual treebank collection}.
\newblock In \emph{Proceedings of the Twelfth Language Resources and Evaluation Conference}, pages 4034--4043, Marseille, France. European Language Resources Association.

\bibitem[{Qi et~al.(2020)Qi, Zhang, Zhang, Bolton, and Manning}]{qi2020stanza}
Peng Qi, Yuhao Zhang, Yuhui Zhang, Jason Bolton, and Christopher~D. Manning. 2020.
\newblock \href {https://nlp.stanford.edu/pubs/qi2020stanza.pdf} {Stanza: A {Python} natural language processing toolkit for many human languages}.
\newblock In \emph{Proceedings of the 58th Annual Meeting of the Association for Computational Linguistics: System Demonstrations}.

\bibitem[{Rush(2020)}]{alex2020torchstruct}
Alexander~M. Rush. 2020.
\newblock \href {https://arxiv.org/abs/2002.00876} {Torch-struct: Deep structured prediction library}.
\newblock \emph{Preprint}, arXiv:2002.00876.

\end{thebibliography}

\end{document}